\documentclass[runningheads]{llncs}
\usepackage[T1]{fontenc}
\usepackage{graphicx}
\newcommand*\samethanks[1][\value{footnote}]{\footnotemark[#1]}
\usepackage[table]{xcolor}

\usepackage{float}

\usepackage{etoolbox}
\apptocmd{\sloppy}{\hbadness 10000\relax}{}{}

\usepackage{tabularray}

\usepackage{graphicx}

\usepackage{hyperref}
\usepackage[T1]{fontenc}

\usepackage{numprint}
\usepackage{newclude}

\usepackage{cleveref}

\usepackage[numbers]{natbib}

\usepackage[nolist]{acronym}
\begin{acronym}
\acro{CNN}{Convolutional Neural Network}
\acro{DL}{Deep Learning}
\acro{DSC}{Dice similarity coefficient}
\acro{GAN}{Generative Adversarial Network}
\acro{ML}{Machine Learning}
\acro{PGT}{Peak Ground Truth}
\acro{RWMP}{Real World Model Performance}
\acro{NN}{Neural Network}
\end{acronym}

\begin{document}
\newcommand{\eac}[1]{\emph{\ac{#1}}}
\newcommand{\eacp}[1]{\emph{\acp{#1}}}
\newcommand{\eacf}[1]{\emph{\acf{#1}}}

%
%
%
%
\title{Approaching \emph{Peak Ground Truth}}

%
%
\author{Florian Kofler\inst{1,2,3,4} \and
Johannes Wahle\inst{14,1} \and
Ivan Ezhov\inst{2,3} \and
Sophia Wagner\inst{1,2} \and
Rami Al-Maskari\inst{2,6} \and
Emilia Gryska\inst{11} \and
Mihail Todorov\inst{6,7} \and \\
Christina Bukas\inst{1} \and
Felix Meissen \inst{2,13} \and
Tingying Peng \inst{1,2} \and
Ali Ertürk \inst{6,7,8,9} \and \\
Daniel Rueckert \inst{2,5} \and
Rolf Heckemann \inst{12} \and
Jan Kirschke \inst{4} \and
Claus Zimmer \inst{4} \and \\
Benedikt Wiestler \inst{4} \thanks{equal contribution}  \and
Bjoern Menze \inst{10} \samethanks \and
Marie Piraud \inst{1} \samethanks
}

\authorrunning{F. Kofler et al.}
%

\institute{
Helmholtz AI, Helmholtz Zentrum München, Germany \and
Department of Informatics, Technical University Munich, Germany \and
TranslaTUM - Central Institute for Translational Cancer Research, Technical University of Munich, Germany \and
Department of Diagnostic and Interventional Neuroradiology, School of Medicine, Klinikum rechts der Isar, Technical University of Munich, Germany \and
Imperial College London, United Kingdom \and
Institute for Tissue Engineering and Regenerative Medicine, Helmholtz Institute Munich (iTERM), Germany \and
Institute for Stroke and dementia research (ISD), University Hospital, LMU Munich, Germany \and
Graduate school of neuroscience (GSN), Munich, Germany \and
Munich cluster for systems neurology (Synergy), Munich, Germany \and
Department of Quantitative Biomedicine, University of Zurich, Switzerland \and
Department of Radiology, Institute of Clinical Sciences, Sahlgrenska Academy, University of Gothenburg, Gothenburg, Sweden \and
Department of Medical Radiation Sciences, University of Gothenburg, Gothenburg, Sweden \and
Chair for AI in Medicine, Klinikum Rechts der Isar, Munich, Germany \and
German Center for Neurodegenerative Diseases (DZNE), Bonn, Germany
}

\maketitle              
\begin{abstract}
Machine learning models are typically evaluated by computing similarity with reference annotations and trained by maximizing similarity with such.
Especially in the biomedical domain, annotations are subjective and suffer from low inter- and intra-rater reliability.
Since annotations only reflect one interpretation of the \emph{real world}, this can lead to sub-optimal predictions even though the model achieves high similarity scores. 
Here, the theoretical concept of \eac{PGT} is introduced.
\eac{PGT} marks the point beyond which an increase in similarity with the \emph{reference annotation} stops translating to better \eac{RWMP}.
Additionally, a quantitative technique to approximate \eac{PGT} by computing inter- and intra-rater reliability is proposed.
Finally, four categories of \eac{PGT}\emph{-aware} strategies to evaluate and improve model performance are reviewed.
\keywords{
machine learning,
deep learning,
annotation,
ground truth,
reference,
segmentation
}

\end{abstract}

\section{Introduction}
Modern \eac{DL} offers countless possibilities for creating models.
Network architecture, optimizer, hyperparameters, and loss function can be arranged in infinite permutations.
Researchers are confronted with the non-trivial task of identifying the best models for their given use case.
A popular way of approaching this, also prominent in classical \eac{ML}, is to compare model outputs with \emph{reference annotations} by computing similarity metrics.
Technical innovations distinguish themselves from established choices by achieving higher similarity scores.
Achieving higher similarity scores is key in marketing innovations and often pivotal for publication decisions.
Several challenges, such as \emph{BraTS} \citep{menze2014multimodal}, \emph{KiTS} \citep{heller2021state}, or \emph{LiTS} \citep{bilic2019liver}, emerged and manifested themselves as a platform for participants to benchmark their algorithms even beyond the bio-medical domain, e.g., \emph{Natural Language Processing} \citep{parra-escartin-etal-2017-ethical}.
Across all these challenges, organizers decorate winners based on their achieved overlap with \emph{reference annotations} \citep{bakas2018identifying}.

\noindent\textbf{Contribution:}
Here, we introduce the theoretical concept of \eac{PGT}.
We define \eac{PGT} as the point beyond which an increase in similarity of a model's output with the annotation will not translate to improved \emph{RWMP}.
Besides proposing means of quantitatively approximating \eac{PGT}, we discuss \eac{PGT}\emph{-aware} strategies to evaluate \eac{ML} model performance.
We illustrate its widespread implications for interpreting \eac{ML} models across and beyond the biomedical domain.

\section{Peak Ground Truth}
\label{sec:peakgt}
The term \eacf{PGT} is inspired by \emph{Peak Oil} \cite{hubbert1949energy}.
In this section, we explain the theoretical foundation of \eac{PGT}, propose means to quantitatively approximate it, and demonstrate its relevance across various \eac{ML} data sets.

\begin{figure*}[ht]
\centering
\label{fig:peak}
\includegraphics[width=1.0\linewidth]{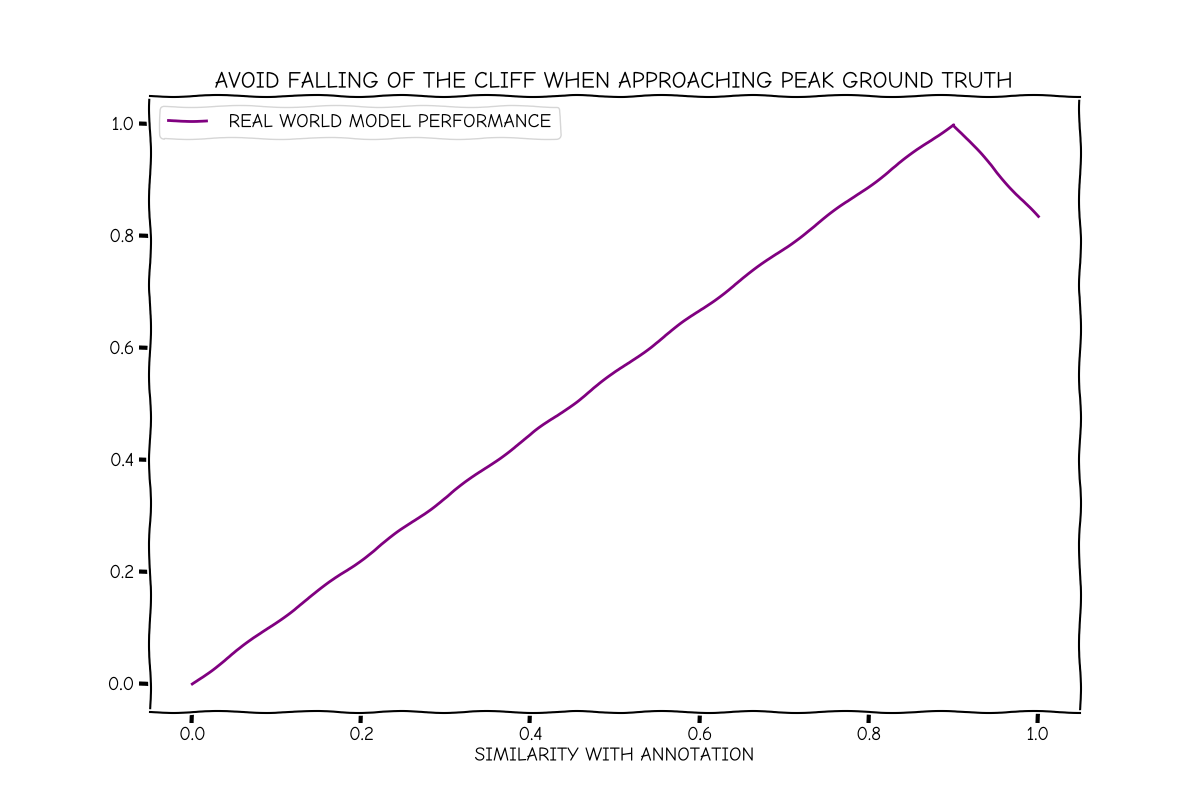}
\caption{
\eacf{PGT} implications for \eac{ML} training.
When employing a similarity metric in the loss function, the network is incentivized to maximize similarity with the \emph{annotation}.
Frequently, similarity with the annotation does not (fully) represent \eacf{RWMP} either/or due to erroneous annotations or poor selection of metrics.
Consequently, increasing similarity only corresponds to increased model performance up to a certain point, as a hypothetical similarity of 100\% would mean reproducing all (potential) errors in the annotation, leading to \emph{falling off the cliff}.
This point is defined as \eacf{PGT}; in this example sketch, it is located around 0.9 on the \emph{x-axis}.
The real location and shape of the curve are defined by the \eacf{ML} problem at hand.
}
\end{figure*}

\subsection{Related Work}
Maier-Hein et al.\ \citep{maier2018rankings} revealed that challenge rankings heavily depend on the choice of evaluation metrics.
Further, Kofler et al.\ \citep{kofler2021we} found that established similarity metrics only moderately correlate with expert perception regarding glioma segmentation of BraTS algorithms.
Similarly, Reinke et al.\ \citep{reinke2021common} illustrate limitations of image processing metrics.
Taha et al.\ \citep{taha2015metrics} analyze the properties of 20 segmentation metrics and provide guidelines for selecting the most appropriate ones for a given task.
Building upon that, Radsch et al.\ \citep{radsch2022labeling} point out that the labeling instructions also matter.
Continuing this trend, in 2020, the \href{http://www.https://qubiq.grand-challenge.org}{QUBIQ Challenge} was hosted for the first time providing participants with multiple reference annotations to reflect variance in the labeling process.

\subsection{Theoretical Concept}
In \eac{ML} research, the terms \emph{ground truth} and \emph{reference labels} are often used synonymously\footnote{In the literature, another synonym for \emph{reference label} is \emph{reference annotation}.
In this manuscript, we use the two terms interchangeably.}. 
To grasp the concept of \eac{PGT}, it is necessary to distinguish them, as reference annotations and the \emph{real world} may not always coincide \citep{Ma_2022_CVPR,Yun_2021_CVPR,pmlr-v119-shankar20c}.

Humans are typically employed in the annotation process for \eac{ML} data set generation.
Even though the biomedical domain typically relies on trained experts, they still produce random and systematic errors due to their human nature.
These annotations are often regarded as the \emph{gold standard} for model evaluation; however, due to their inaccuracies, they should not be regarded as \emph{ground truth}.

Model performance is commonly evaluated by measuring the similarity between model outputs and \emph{reference annotations}.
A \eac{PGT} situation emerges when this measurement does not (fully) represent \eacf{RWMP}.
Such a discrepancy can manifest itself  from inaccuracies in the \emph{reference annotations} and/or a poor choice of metrics.
Now, \eac{PGT} represents the point after which an increase in similarity with the \emph{reference labels} leads to a decrease in \eac{RWMP} while the model (randomly) fits the errors in the \emph{reference annotation}.
The concept and its implications for \eac{ML} training are illustrated in \Cref{fig:peak}.

Importantly, \eac{PGT} needs to be distinguished from the \emph{generalization error}; even though provided an accurate choice of evaluation metric(s),  the \emph{generalization error's} minimum is expected to localize around \eac{PGT}. 
Exceeding \eac{PGT} needs to be distinguished from \emph{overfitting} as well.
While \emph{overfitting} refers to the training data, \eac{PGT} refers to achieving a random fit with the (untrained) test data.

\subsection{Quantitative Approximation}
Where is \eac{PGT} located?
In other words, at which point should one stop optimizing similarity metrics?

The location of \eac{PGT} is determined by two factors.
First, the similarity metrics' ability to capture the \emph{real-world} problem and second, the validity of our annotation entities.
Thus, we propose the following multi-step process to narrow down the location of \eac{PGT}:

First, the \eac{ML} problem needs to be formalized to properly represent the \emph{real-world} problem by selecting an appropriate metric, cf. \Cref{sec:metric_selection}.
Second, we determine the validity of the annotation.
As usual, there is no direct measure for validity; we need to approximate it.
Therefore, we determine the inter-rater reliability to serve as the lower- and the intra-rater reliability to serve as the upper bound.
Now, the \emph{real} \eac{PGT} with high likelihood lies between these two values.
Here, it is important to note that systematic measurement errors (\emph{lack of validity}) will lead to overestimation of the upper and lower bound.
Consequently, one can escape the \eac{PGT} problem by combining perfect annotations with perfect similarity metric(s).

In a practical example, high inter-rater reliability suggests there is strong agreement between annotators.
Consequently, we expect \eac{PGT} at a higher similarity between model output and \emph{reference}.
On the other hand, high intra-rater reliability indicates strong consistency of the annotation entity.
Exceeding the intra-rater reliability indicates the model started to (randomly) fit the annotation noise.

\subsection{Practical examples}
Annotation inaccuracies are common in \eac{ML} data sets, such as \emph{COCO, ImageNet, RTE-1, BraTS} or \emph{ChestX-ray8} \citep{Ma_2022_CVPR,Yun_2021_CVPR,pmlr-v119-shankar20c,beyer2020we,klebanov2010some,kofler2021we,wang2017chestx}.
Bio-medical imaging techniques often contain modality-specific artifacts overshadowing the annotation process.

\begin{figure}[ht]
\includegraphics[width=1.0\linewidth]{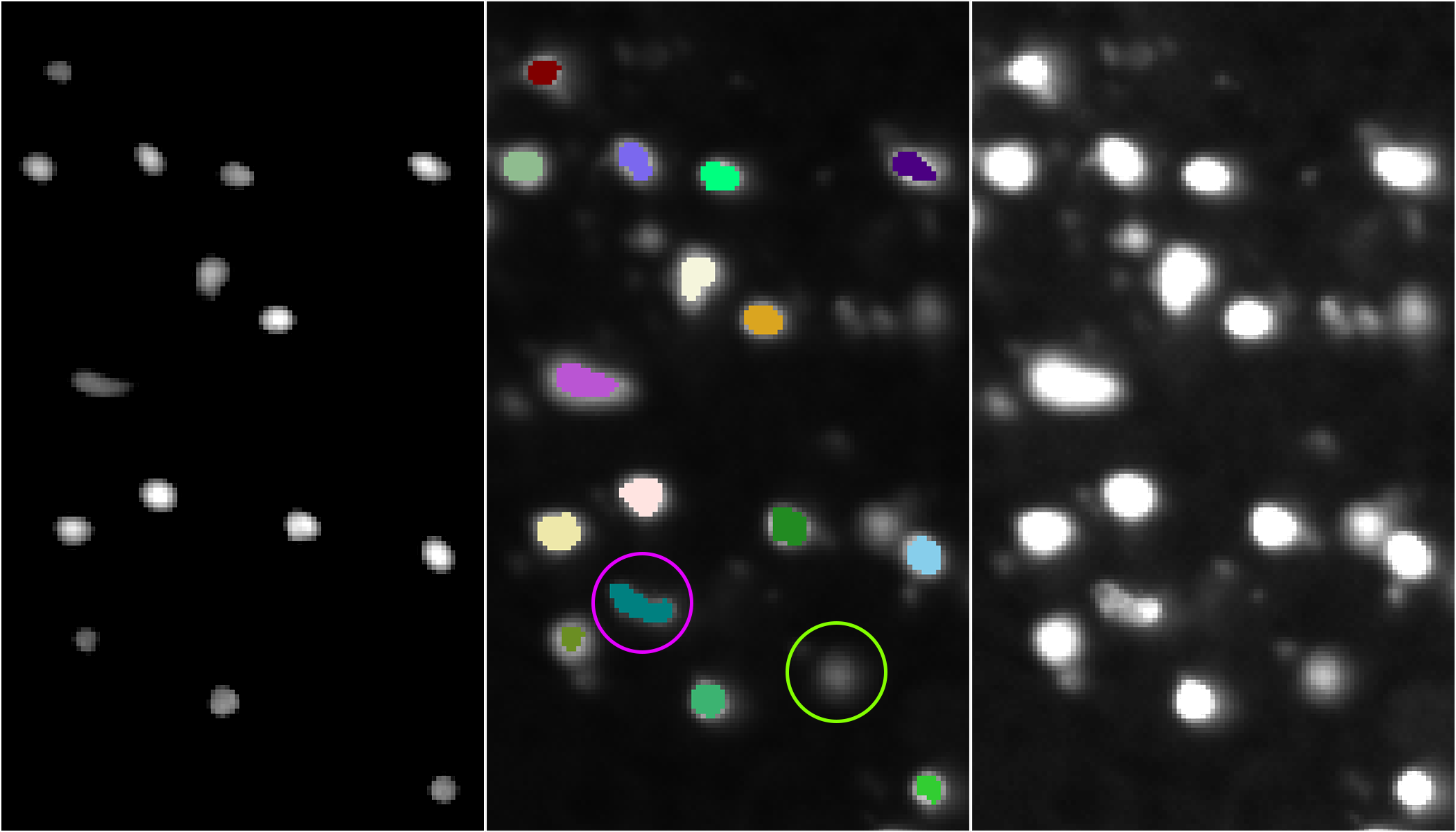}
\caption{
Dependency on thresholding in light sheet fluorescence microscopy -- slices depicting human neurons from the public \emph{SHANEL} dataset \citep{zhao2020cellular} at different thresholds.
\emph{Left}: the neurons at a low threshold.
\emph{Middle}: the neurons at a medium threshold overlayed with the annotation of the expert neurobiologist.
Colors represent the instances (neurons).
\emph{Right}: the neurons at a high threshold.
Depending on the arbitrary intensity threshold setting, the objects appear in varying sizes.
This effect is particularly pronounced in light sheet microscopy, as the illumination is perpendicular to the objective.
Consequently, the annotations can appear on the full spectrum from \emph{over-} to \emph{under-segmented}.
Notably, only one of the neurons encircled in \emph{magenta} and \emph{lime} is annotated, while neither appears in the \emph{left} image.
This illustrates the subjectivity of human-curated \emph{labels} and thus the discrepancy between \emph{annotations} and \emph{ground truth}.
}
\label{fig:micros}
\end{figure}

These are particularly pronounced in light microscopy.
In fluorescence microscopy, objects appear smaller or larger in size depending on the setting of the threshold intensity; cf. \Cref{fig:micros}.
Consequently, the volumes of the resulting annotations are a function of the threshold setting.
Furthermore, thresholding stands in complex interaction with biological parameters, such as the quality of the applied clearing, the tissue's autofluorescence, or fluorophore amplification \citep{cai2019panoptic}.

It is important to note that comparable issues arise for other imaging techniques.
For instance, \emph{Magnetic Resonance} or \emph{Computed Tomography} images frequently suffer from artifacts induced due to motion, field inhomogeneity, etc.
This should be taken into account when interpreting volumetric measures, such as the popular \eac{DSC}.

\section{Mitigation Strategies}
\label{sec:mitigation}
In this section, we review potential mitigation strategies to account for \eac{PGT}.
Broadly these can be classified into four categories:

\subsection{Improving Annotation Quality}
One mitigation strategy consists in improving the annotation quality to increase the \emph{annotation's} validity.
This allows shifting \ac{PGT} to the \emph{"right"}, compare \Cref{fig:peak}.
Therefore, the decline in \eac{RWMP} can be expected to occur at a higher level of similarity.
Here, a prominent approach marks the ensembling of annotation entities, for example, by employing consensus voting \citep{klebanov2010some,MaierHein2016, qing-etal-2014-empirical,yang2022neural}.
Interactive \emph{Human in the loop} approaches represent an active field of research trying to develop means to leverage the complementary potentials of machines and humans  \citep{Mosqueira-Rey2022,BUDD2021102062}.
Rädsch et al.\ demonstrate that annotation instructions should also be carefully formulated \citep{radsch2022labeling} to avoid erroneous expert annotations.

\subsection{Embracing Annotation Noise}
Another way forward is to recognize the inevitability of annotation errors \citep{Frenay2014}.
Jiang et al.\ propose a learning scheme to benchmark and alleviate the impact of label noise \citep{jiang2020beyond}.
Alternatively, Bootkrajang adds a term to explicitly model the probability of misclassification \citep{BOOTKRAJANG201661}.
Kohl et al. introduce a probabilistic method to generate plausible segmentation maps to embrace ambiguities in the annotation \citep{kohl2018probabilistic}.

\subsection{Optimizing Metric Choice}
\label{sec:metric_selection}
Apart from annotation inaccuracies, \eac{PGT} significantly depends on the metrics' ability to capture \eac{RWMP}.
Maier-Hein, Reinke, and others \citep{reinke2021common, maier2022metrics, reinke2022metrics, REINKE2021710} provide guidelines for tailoring the metric selection around the underlying \emph{real-world} problem.
In the absence of situation-specific similarity metrics, a popular mitigation strategy represents optimizing multiple metrics at the same time.
Challenge organizers also frequently adopt this strategy \citep{maier2018rankings}.
Similarly, for the training of semantic segmentation models, compound loss functions, such as combining soft Dice loss with binary cross-entropy loss,  have become increasingly popular \citep{isensee2021nnu,kofler2021we}.

\subsection{Choosing Annotation Independent Metrics}
Another potential mitigation strategy is the introduction of performance measures that are independent of reference labels.
Due to the lack of reference annotations, the evaluation of \eacp{GAN} is far more developed in this regard.
Borji reviews several quantitative and qualitative measures for evaluating \eacp{GAN} \citep{borji2019pros,BORJI2022103329}.
One of these measures, also employed for measuring the quality of segmentation models, are human expert ratings \citep{kofler2021we}.
Building upon this, Kofler et al.\ \citep{kofler2022deep} demonstrate that surrogate models can approximate human expert ratings.
Further, the developing field of perceptual metrics promises to generate new label-independent metrics.
For instance, Bhardwaj et al.\ \citep{NEURIPS2020_Bhardwaj} propose \emph{PIM}, a perceptual metric grounded in information theory.
Additionally, Corneanu et al.\ \citep{corneanu2020computing} propose persistent topology measures to predict model performance on unseen samples.
Quantifying model performance through a proxy downstream task is another reference-independent strategy.
However, such a pipeline has to be carefully designed to avoid introduction of further evaluation noise.

\section{Discussion}
In this work, we introduce the theoretical concept of \eacf{PGT}.
Further, we suggest a quantitative approximation based on inter- and intra-rater reliability.
We point out difficulties in human-curated annotation and illustrate it for the example of instance-segmentation in light-sheet microscopy.
In addition, we provide \eac{PGT}\emph{-aware} strategies to evaluate model performance.

\noindent\textbf{Limitations:}
Even though our method can efficiently approximate \eac{PGT}, exact quantification requires an accurate measure of \eacf{RWMP} and extensive experimentation.
Likewise, systematic measurement errors may lead to overestimations of the \eac{PGT} approximation method.
Our quantification approach relies on multiple annotation entities.
Obtaining these can be expensive, especially in the bio-medical domain, which typically relies on scarce human experts.

\noindent\textbf{Outlook:}
In current \eac{ML} research, it is common to market technical innovations by demonstrating small increases in similarity with reference annotations.
However, beyond \eac{PGT}, this does not necessarily translate to improved \eac{RWMP} and arguably just represents randomly fitting imperfections of the reference annotations.
We hope our work will contribute to inciting a discussion over these practices and help the field move forward.

\vspace{\baselineskip}


\section*{Acknowledgement}
\noindent BM, BW, and FK are supported through the SFB 824, subproject B12.

\noindent Supported by Deutsche Forschungsgemeinschaft (DFG) through TUM International Graduate School of Science and Engineering (IGSSE), GSC 81.

\noindent IE is supported by the Translational Brain Imaging Training Network (TRABIT) under the European Union's `Horizon 2020' research \& innovation program (Grant agreement ID: 765148).

\noindent IE is funded by DComEX (Grant agreement ID: 956201).

\noindent SJW was supported by the Helmholtz Association under the joint research school “Munich School for Data Science -- MUDS” and the Add-on Fellowship of the Joachim Herz Foundation.

\noindent Supported by Anna Valentina Lioba Eleonora Claire Javid Mamasani.

\noindent With the support of the Technical University of Munich – Institute for Advanced Study, funded by the German Excellence Initiative.

\noindent JK has received Grants from the ERC, DFG, BMBF and is Co-Founder of Bonescreen GmbH.

\noindent BM acknowledges support by the Helmut Horten Foundation.

\vspace{15mm}
%
%
%
\let\clearpage\relax
\bibliographystyle{splncs04}
\bibliography{references}

\begin{thebibliography}{10}
\providecommand{\url}[1]{\texttt{#1}}
\providecommand{\urlprefix}{URL }
\providecommand{\doi}[1]{https://doi.org/#1}

\bibitem{bakas2018identifying}
Bakas, S., Reyes, M., et~al.: Identifying the best machine learning algorithms
  for brain tumor segmentation, progression assessment, and overall survival
  prediction in the brats challenge. arXiv preprint arXiv:1811.02629  (2018)

\bibitem{beyer2020we}
Beyer, L., H{\'e}naff, O.J., et~al.: Are we done with imagenet? arXiv preprint
  arXiv:2006.07159  (2020)

\bibitem{NEURIPS2020_Bhardwaj}
Bhardwaj, S., Fischer, I., et~al.: An unsupervised information-theoretic
  perceptual quality metric. In: Larochelle, H., Ranzato, M., Hadsell, R.,
  Balcan, M., Lin, H. (eds.) Advances in Neural Information Processing Systems.
  vol.~33, pp. 13--24. Curran Associates, Inc. (2020),
  \url{https://proceedings.neurips.cc/paper/2020/file/00482b9bed15a272730fcb590ffebddd-Paper.pdf}

\bibitem{bilic2019liver}
Bilic, P., Christ, P.F., et~al.: The liver tumor segmentation benchmark (lits).
  arXiv preprint arXiv:1901.04056  (2019)

\bibitem{BOOTKRAJANG201661}
Bootkrajang, J.: A generalised label noise model for classification in the
  presence of annotation errors. Neurocomputing  \textbf{192},  61--71 (2016).
  \doi{https://doi.org/10.1016/j.neucom.2015.12.106},
  \url{https://www.sciencedirect.com/science/article/pii/S0925231216002551},
  advances in artificial neural networks, machine learning and computational
  intelligence

\bibitem{borji2019pros}
Borji, A.: Pros and cons of gan evaluation measures. Computer Vision and Image
  Understanding  \textbf{179},  41--65 (2019)

\bibitem{BORJI2022103329}
Borji, A.: Pros and cons of gan evaluation measures: New developments. Computer
  Vision and Image Understanding  \textbf{215},  103329 (2022).
  \doi{https://doi.org/10.1016/j.cviu.2021.103329},
  \url{https://www.sciencedirect.com/science/article/pii/S1077314221001685}

\bibitem{BUDD2021102062}
Budd, S., Robinson, E.C., Kainz, B.: A survey on active learning and
  human-in-the-loop deep learning for medical image analysis. Medical Image
  Analysis  \textbf{71},  102062 (2021).
  \doi{https://doi.org/10.1016/j.media.2021.102062},
  \url{https://www.sciencedirect.com/science/article/pii/S1361841521001080}

\bibitem{cai2019panoptic}
Cai, R., Pan, C., et~al.: Panoptic imaging of transparent mice reveals
  whole-body neuronal projections and skull--meninges connections. Nature
  neuroscience  \textbf{22}(2),  317--327 (2019)

\bibitem{corneanu2020computing}
Corneanu, C.A., Escalera, S., et~al.: Computing the testing error without a
  testing set. In: Proceedings of the IEEE/CVF Conference on Computer Vision
  and Pattern Recognition. pp. 2677--2685 (2020)

\bibitem{Frenay2014}
Frenay, B., Verleysen, M.: Classification in the presence of label noise: A
  survey. IEEE Transactions on Neural Networks and Learning Systems
  \textbf{25}(5),  845--869 (2014). \doi{10.1109/TNNLS.2013.2292894}

\bibitem{heller2021state}
Heller, N., Isensee, F., et~al.: The state of the art in kidney and kidney
  tumor segmentation in contrast-enhanced ct imaging: Results of the kits19
  challenge. Medical image analysis  \textbf{67},  101821 (2021)

\bibitem{hubbert1949energy}
Hubbert, M.K.: Energy from fossil fuels. Science  \textbf{109}(2823),  103--109
  (1949)

\bibitem{isensee2021nnu}
Isensee, F., Jaeger, P.F., et~al.: nnu-net: a self-configuring method for deep
  learning-based biomedical image segmentation. Nature methods  \textbf{18}(2),
   203--211 (2021)

\bibitem{jiang2020beyond}
Jiang, L., Huang, D., et~al.: Beyond synthetic noise: Deep learning on
  controlled noisy labels. In: International Conference on Machine Learning.
  pp. 4804--4815. PMLR (2020)

\bibitem{klebanov2010some}
Klebanov, B.B., Beigman, E.: Some empirical evidence for annotation noise in a
  benchmarked dataset. In: Human Language Technologies: The 2010 Annual
  Conference of the North American Chapter of the Association for Computational
  Linguistics. pp. 438--446 (2010)

\bibitem{kofler2021we}
Kofler, F., Ezhov, I., et~al.: Are we using appropriate segmentation metrics?
  identifying correlates of human expert perception for cnn training beyond
  rolling the dice coefficient. arXiv preprint arXiv:2103.06205  (2021)

\bibitem{kofler2022deep}
Kofler, F., Ezhov, I., et~al.: Deep quality estimation: Creating surrogate
  models for human quality ratings. arXiv preprint arXiv:2205.10355  (2022)

\bibitem{kohl2018probabilistic}
Kohl, S., Romera-Paredes, B., et~al.: A probabilistic u-net for segmentation of
  ambiguous images. Advances in neural information processing systems
  \textbf{31} (2018)

\bibitem{Ma_2022_CVPR}
Ma, J., Ushiku, Y., Sagara, M.: The effect of improving annotation quality on
  object detection datasets: A preliminary study. In: Proceedings of the
  IEEE/CVF Conference on Computer Vision and Pattern Recognition (CVPR)
  Workshops. pp. 4850--4859 (June 2022)

\bibitem{MaierHein2016}
Maier-Hein, L., Ross, T., et~al.: Crowd-algorithm collaboration for large-scale
  endoscopic image annotation with confidence. In: Ourselin, S., Joskowicz, L.,
  Sabuncu, M.R., Unal, G., Wells, W. (eds.) Medical Image Computing and
  Computer-Assisted Intervention -- MICCAI 2016. pp. 616--623. Springer
  International Publishing, Cham (2016)

\bibitem{maier2018rankings}
Maier-Hein, L., Eisenmann, M., et~al.: Why rankings of biomedical image
  analysis competitions should be interpreted with care. Nature communications
  \textbf{9}(1),  1--13 (2018)

\bibitem{maier2022metrics}
Maier-Hein, L., Reinke, A., et~al.: Metrics reloaded: Pitfalls and
  recommendations for image analysis validation. arXiv preprint
  arXiv:2206.01653  (2022)

\bibitem{menze2014multimodal}
Menze, B.H., Jakab, A., et~al.: The multimodal brain tumor image segmentation
  benchmark (brats). IEEE transactions on medical imaging  \textbf{34}(10),
  1993--2024 (2014)

\bibitem{Mosqueira-Rey2022}
Mosqueira-Rey, E., Hern{\'a}ndez-Pereira, E., et~al.: Human-in-the-loop machine
  learning: a state of the art. Artificial Intelligence Review  (2022).
  \doi{10.1007/s10462-022-10246-w},
  \url{https://doi.org/10.1007/s10462-022-10246-w}

\bibitem{parra-escartin-etal-2017-ethical}
Parra~Escart{\'\i}n, C., Reijers, W., et~al.: Ethical considerations in {NLP}
  shared tasks. In: Proceedings of the First {ACL} Workshop on Ethics in
  Natural Language Processing. pp. 66--73. Association for Computational
  Linguistics, Valencia, Spain (Apr 2017). \doi{10.18653/v1/W17-1608},
  \url{https://aclanthology.org/W17-1608}

\bibitem{qing-etal-2014-empirical}
Qing, C., Endriss, U., et~al.: Empirical analysis of aggregation methods for
  collective annotation. In: Proceedings of {COLING} 2014, the 25th
  International Conference on Computational Linguistics: Technical Papers. pp.
  1533--1542. Dublin City University and Association for Computational
  Linguistics, Dublin, Ireland (Aug 2014),
  \url{https://aclanthology.org/C14-1145}

\bibitem{radsch2022labeling}
R{\"a}dsch, T., Reinke, A., et~al.: Labeling instructions matter in biomedical
  image analysis. arXiv preprint arXiv:2207.09899  (2022)

\bibitem{reinke2021common}
Reinke, A., Eisenmann, M., et~al.: Common limitations of image processing
  metrics: A picture story. arXiv preprint arXiv:2104.05642  (2021)

\bibitem{reinke2022metrics}
Reinke, A., Maier-Hein, L., et~al.: Metrics reloaded-a new recommendation
  framework for biomedical image analysis validation. In: Medical Imaging with
  Deep Learning (2022)

\bibitem{REINKE2021710}
Reinke, A., Tizabi, M.D., et~al.: Common pitfalls and recommendations for grand
  challenges in medical artificial intelligence. European Urology Focus
  \textbf{7}(4),  710--712 (2021).
  \doi{https://doi.org/10.1016/j.euf.2021.05.008},
  \url{https://www.sciencedirect.com/science/article/pii/S2405456921001607}

\bibitem{pmlr-v119-shankar20c}
Shankar, V., Roelofs, R., et~al.: Evaluating machine accuracy on {I}mage{N}et.
  In: III, H.D., Singh, A. (eds.) Proceedings of the 37th International
  Conference on Machine Learning. Proceedings of Machine Learning Research,
  vol.~119, pp. 8634--8644. PMLR (13--18 Jul 2020),
  \url{https://proceedings.mlr.press/v119/shankar20c.html}

\bibitem{taha2015metrics}
Taha, A.A., Hanbury, A.: Metrics for evaluating 3d medical image segmentation:
  analysis, selection, and tool. BMC medical imaging  \textbf{15}(1),  1--28
  (2015)

\bibitem{wang2017chestx}
Wang, X., Peng, Y., et~al.: Chestx-ray8: Hospital-scale chest x-ray database
  and benchmarks on weakly-supervised classification and localization of common
  thorax diseases. In: Proceedings of the IEEE conference on computer vision
  and pattern recognition. pp. 2097--2106 (2017)

\bibitem{yang2022neural}
Yang, J., Shi, R., et~al.: Neural annotation refinement: Development of a new
  3d dataset for adrenal gland analysis. In: International Conference on
  Medical Image Computing and Computer-Assisted Intervention. pp. 503--513.
  Springer (2022)

\bibitem{Yun_2021_CVPR}
Yun, S., Oh, S.J., et~al.: Re-labeling imagenet: From single to multi-labels,
  from global to localized labels. In: Proceedings of the IEEE/CVF Conference
  on Computer Vision and Pattern Recognition (CVPR). pp. 2340--2350 (June 2021)

\bibitem{zhao2020cellular}
Zhao, S., Todorov, M.I., et~al.: Cellular and molecular probing of intact human
  organs. Cell  \textbf{180}(4),  796--812 (2020)

\end{thebibliography}

\end{document}